\documentclass[runningheads]{llncs}
\usepackage{graphicx}
\usepackage{subfiles}
\usepackage{threeparttable}
\usepackage{subcaption}
\usepackage{amsmath}

\begin{document}
\title{Online panoptic 3D reconstruction as a Linear Assignment Problem}
%
%
\author{Leevi Raivio\inst{1}\orcidID{0000-0002-6902-5201} \and
Esa Rahtu\inst{1}\orcidID{0000-0001-8767-0864}}
\authorrunning{L. Raivio and E. Rahtu}
%
\institute{Tampere University, Korkeakoulunkatu 7, 33720 Tampere, Finland \\ \email{firstname.lastname@tuni.fi}}
\maketitle              
\begin{abstract}

Real-time holistic scene understanding would allow machines to interpret their surrounding in a much more detailed manner than is currently possible. While panoptic image segmentation methods have brought image segmentation closer to this goal, this information has to be described relative to the 3D environment for the machine to be able to utilise it effectively. In this paper, we investigate methods for sequentially reconstructing static environments from panoptic image segmentations in 3D. We specifically target real-time operation: the algorithm must process data strictly online and be able to run at relatively fast frame rates. Additionally, the method should be scalable for environments large enough for practical applications. By applying a simple but powerful data-association algorithm, we outperform earlier similar works when operating purely online. Our method is also capable of reaching frame-rates high enough for real-time applications and is scalable to larger environments as well. Source code and further demonstrations are released to the public at: \url{https://tutvision.github.io/Online-Panoptic-3D/}

\end{abstract}
\section{Introduction}
    \label{sec:intro}
    
Panoptic segmentation \cite{panoptic_segmentation} is a recent computer vision topic, which combines the tasks of semantic segmentation -- assign a class label to each pixel -- and instance segmentation -- detect, classify and segment each object instance. The objective is to segment and classify both \textit{stuff} -- amorphous, unquantifiable areas of the image like floor, buildings and roads -- and \textit{things} -- quantifiable objects. Many potential applications -- \textit{e.g.} in the fields of context-aware augmented reality \cite{panopticfusion}, autonomous driving \cite{real_time_panoptic} and robotics \cite{interactive_3d_scenes} -- require rich semantic information in real time from the environment. For instance, real-time semantic knowledge of objects around a robot would allow it to interact with it's environment at a higher level of autonomy and utilise the information for more robust localisation.

While panoptic image segmentation has been researched quite extensively \cite{panopticfpn,seamless,panoptic_deeplab,efficientps,real_time_panoptic}, panoptic 3D reconstruction has not been studied as much. Segmented images alone are often not sufficient: in many applications the segments 3D location relative to the actor needs to be known as well. The segmentation of 3D reconstructions has gained quite a lot of attention recently \cite{scannet,s3dis,paris_lille_3d}, but many works assume offline processing -- \textit{i.e.} that all the data is available simultaneously rather than sequentially -- which rules out many real-time applications. 


\begin{figure}[t]
\centering
\begin{subfigure}[b]{0.32\textwidth}
        \centering
        \includegraphics[width=\textwidth]{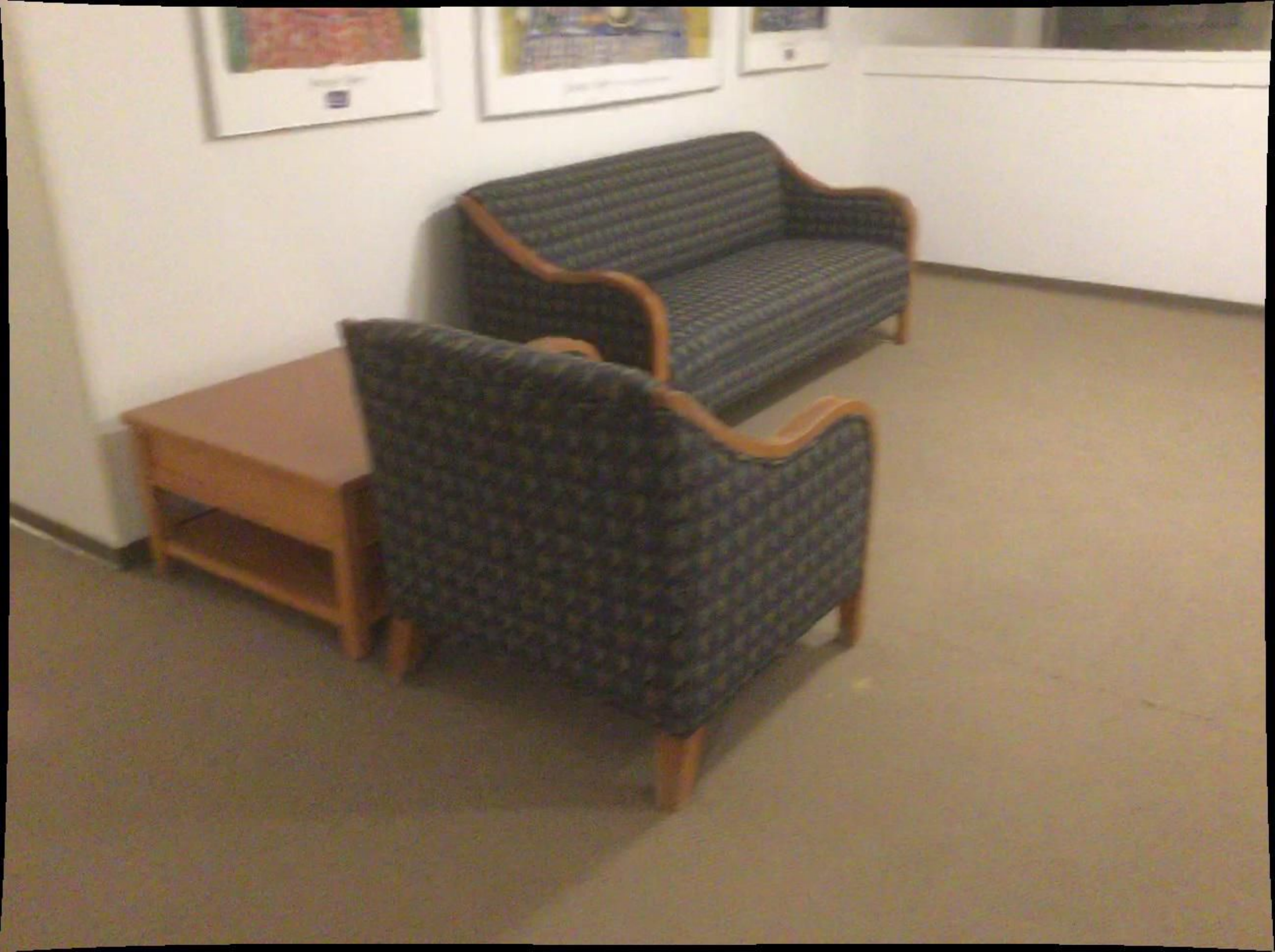}
        \caption[]%
        {{\small Original image}}    
        \label{fig:intro_orig}
\end{subfigure}
\hfill
\begin{subfigure}[b]{0.32\textwidth}
        \centering
        \includegraphics[width=\textwidth]{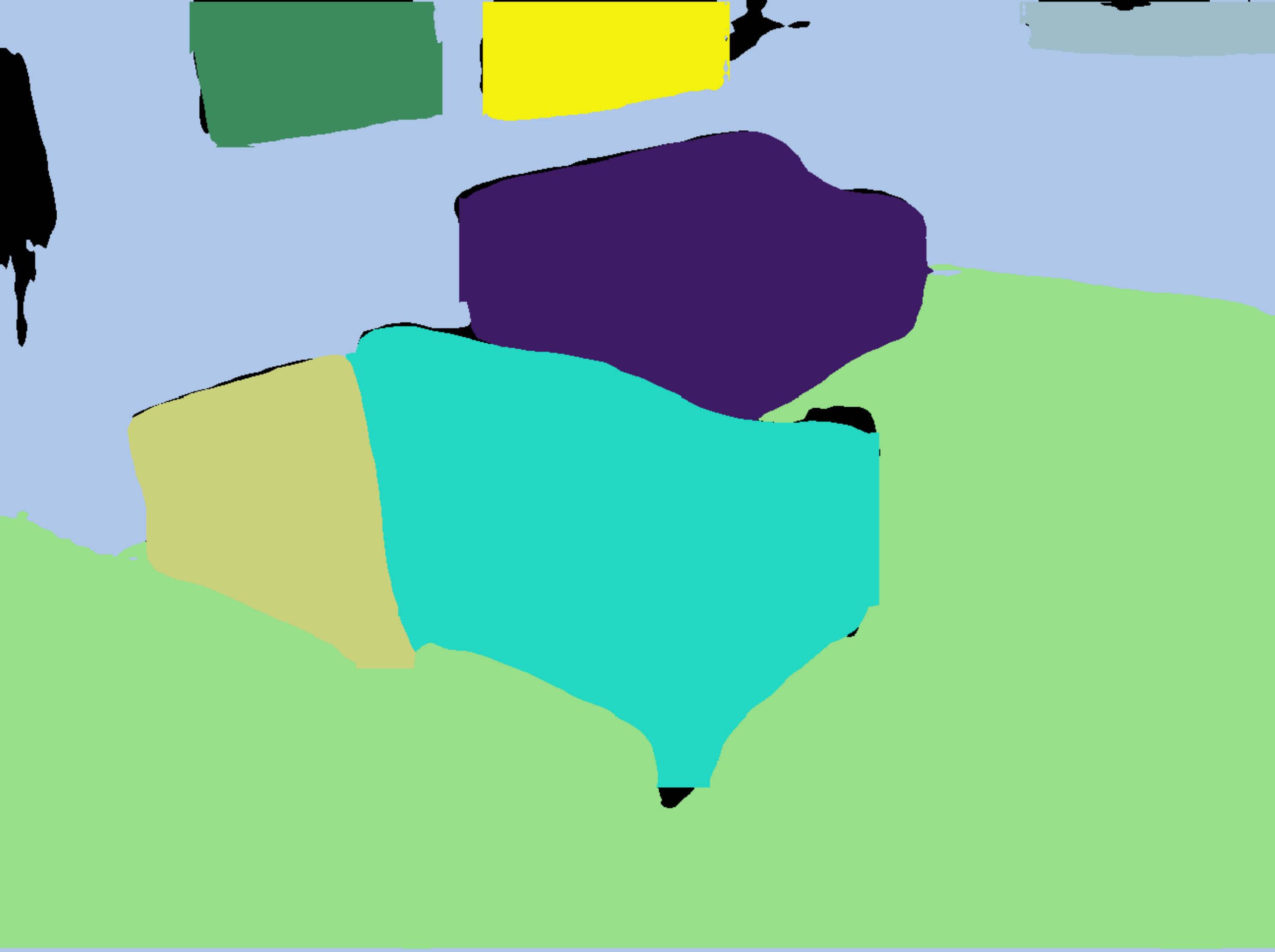}
        \caption[]%
        {{\small Segmented image}}    
        \label{fig:intro_2d_seg}
\end{subfigure}
\hfill
\begin{subfigure}[b]{0.32\textwidth}
        \centering
        \includegraphics[width=\textwidth]{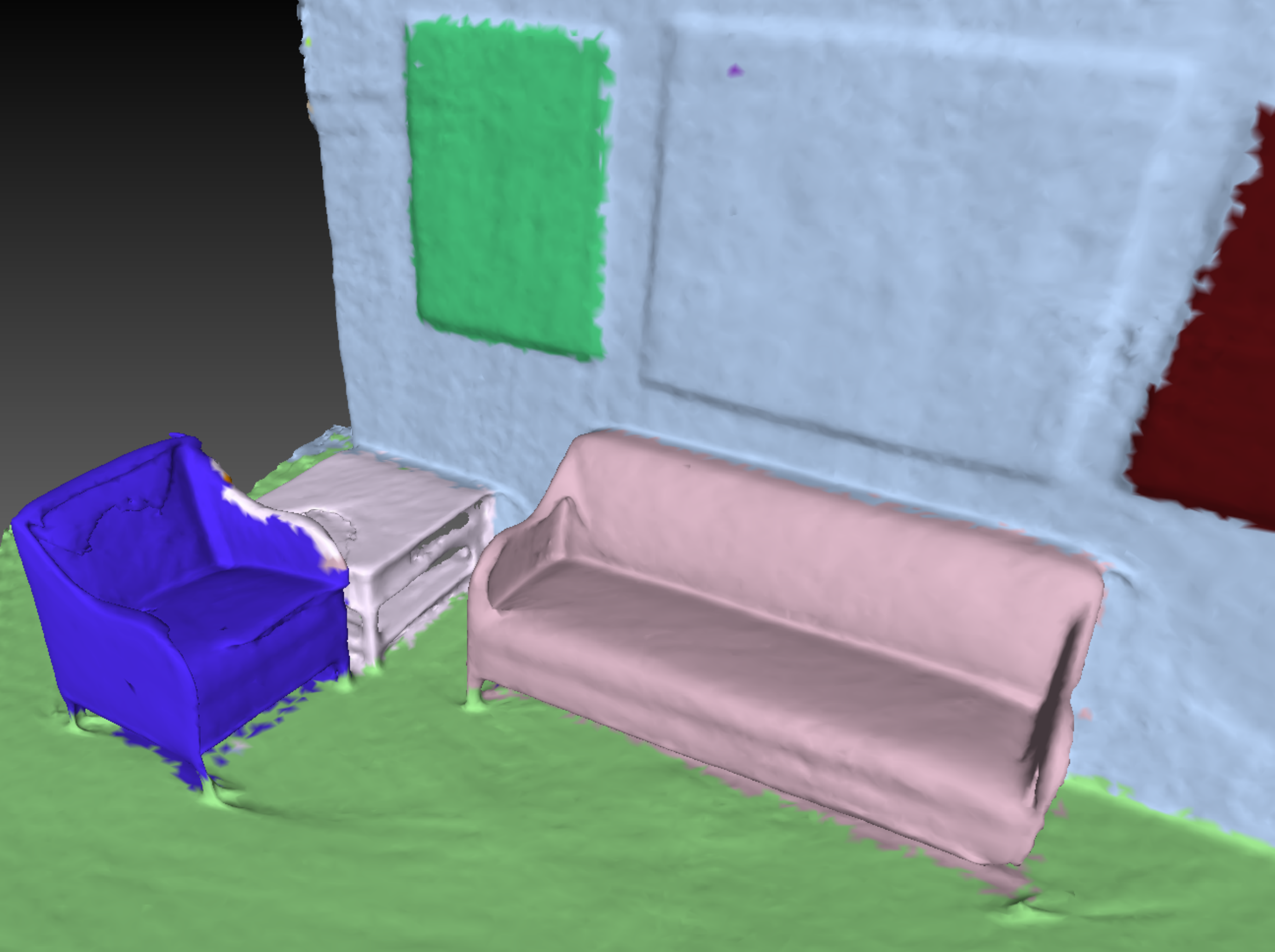}
        \caption[]%
        {{\small 3D reconstruction}}    
        \label{fig:ntro_3d_seg}
\end{subfigure}
\caption[]{An example of online panoptic 3D reconstruction. RGB-D Images (a) are segmented (b) and integrated sequentially into a panoptic 3D reconstruction (c)}
\label{fig:intro_example}
\end{figure}




\subsubsection{Contributions and scope} This work is focused on online panoptic 3D reconstruction of static scenes from sequences of RGB-D images or point clouds. Segmentation of completed 3D reconstructions and other ways of achieving the same result offline are considered out of scope. Since the focus is on what happens after panoptic segmentation, the related segmentation methods are only discussed briefly. The main contributions in this article are as follow:

\begin{itemize}
    \item[\textbullet] We revisit the method introduced with PanopticFusion \cite{panopticfusion}, explore some unanswered aspects and provide an open-source implementation with the updated method,
    \item[\textbullet] formulate online panoptic 3D data-association as a Linear Assignment Problem (LAP), separate from segmentation and reconstruction,
    \item[\textbullet] provide a simple yet effective baseline for real-time operation, with TSDF integration and optimal LAP solution in relation to association likelihood.
\end{itemize}

\subsubsection{Organisation of the rest of the article}

Background and a brief introduction to the applied methods are provided in Section \ref{sec:background}. Section \ref{sec:methods} introduces our proposed method, which is evaluated in Section \ref{sec:evaluation}. Finally, Section \ref{sec:conclusion} concludes the paper, adding some final remarks on the system's applications and future possibilities.

\section{Related work}
    \label{sec:background}

PanopticFusion \cite{panopticfusion} is the first work to propose online integration of panoptic image segmentations to a 3D reconstruction. They integrate point clouds generated from segmented images to a TSDF voxel volume \cite{tsdf,voxblox} by greedily matching detected segments with the reconstruction and regulating each voxel's corresponding instance with a weighting function. Semantic labels are inferred in a bayesian manner based on confidence scores provided by the segmentation model. They also apply a Conditional Random Field (CRF) to regularise the reconstruction, improving results significantly. Voxblox++ \cite{voxblox++} -- introduced later the same year -- is a similar approach that also integrates segmented RGB-D images into a TSDF volume. It leverages geometric segmentation of depth images to improve instance segmentation accuracy. Both geometric and semantic segments are used to compute a pair-wise weight, which is used to greedily match them with segments in the reconstruction. Because of the geometric segmentation, the method allows segmentation of objects with no known semantic class in addition to objects recognised by the instance segmentation model. 

Recently, \cite{interactive_3d_scenes} built upon the idea of Voxblox++. They apply Voxblox++ for 3D instance integration, with two small but effective modifications: the pair-wise weight is replaced by a triplet weight that also takes semantic labels into account in the fusion, and -- in addition to geometric segments -- instance segments are fused if they overlap by a significant amount. The article introduces a method for searching and aligning CAD models to reconstructed objects based on geometry and semantic class, as well as geometrical and physical rules. With the CAD models, a contact graph and interactive virtual scene are reconstructed to allow a robot to simulate its interaction with the environment. SceneGraphFusion \cite{scenegraphfusion} is another approach that forms a scene graph online from a stream of RGB-D images, but unlike the above-mentioned approach, it generates the graph with a deep neural network, after which the panoptic labels for geometrically segmented portions of the 3D reconstruction are produced a side product.

Panoptic-MOPE \cite{panoptic_mope} is another recent approach, which integrates sequences of RGB-D images into a surfel reconstruction. Unlike other mentioned approaches -- which assume the camera pose either known or estimated elsewhere -- it also tracks camera movements based on geometric-, appearance- and semantic cues. The method also applies a novel RGB-D panoptic segmentation model. Although it is only tested on room-sized environments, the authors claim it could be scaled to larger environments as well.

\section{Methods}
    \label{sec:methods}
    
\begin{figure}[t]

    \centering
    \includegraphics[width=\textwidth]{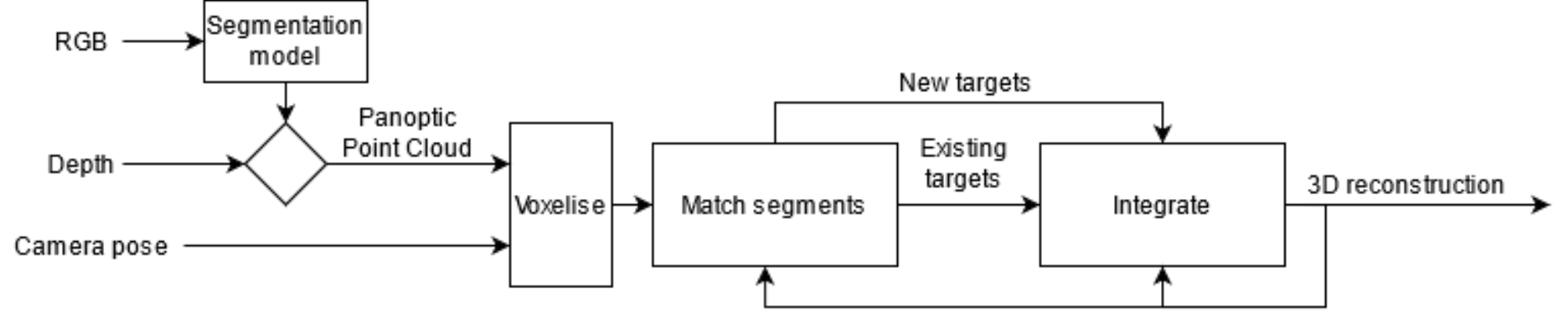}
    \caption[]{A flow diagram of the reconstruction pipeline}
\label{fig:methods_intro_diagram}
\end{figure}

Figure \ref{fig:methods_intro_diagram} depicts a flow diagram of our panoptic 3D reconstruction pipeline. First, instance IDs and semantic classes are acquired from RGB images with a panoptic segmentation model. The IDs and classes are combined with depth information to form a panoptic point cloud, which is transformed to a global coordinate frame with camera pose information and quantised to a voxel grid. Voxel clusters corresponding to detections are matched with ones found in the current reconstruction. Afterwards, the new voxels are integrated into the volume. If there are no matches for some of the detected segments, new targets are generated accordingly.

Similar to \cite{panopticfusion}, \cite{voxblox++} and \cite{interactive_3d_scenes}, we apply the popular Voxblox ROS library \cite{voxblox} to produce a TSDF voxel volume online. Our panoptic tracking system is built alongside the TSDF integrator, separate from the segmentation model and the integrator itself. Although our implementation isn't highly optimised, some parallelisation has been utilised to reach competitive speeds. We apply EfficientPS \cite{efficientps} for segmentation because of its efficient design and impressive performance on benchmark datasets.

    \subsection{Instance association as a Linear Assignment Problem}
        \label{subsec:association}
        



The most essential part of integrating 2D panoptic segmentations to a 3D reconstruction is the association of detected segments with both temporally and spatially consistent labels. While classes determined as \textit{stuff} -- amorphous, unquantifiable regions like sky, road or floor -- can always simply be associated with the same persistent ID, the association of \textit{things} -- quantifiable objects -- is not as trivial. The output of an image segmentation model is only consistent within the single frame: the same instance label is guaranteed to point to the same object only in the image it originates from. On the other hand, while video segmentation methods like \cite{video_panoptic_segmentation} can be applied to maintain temporal consistency of instance labels in the current camera view, they can not necessarily hold spatial consistency throughout the sequence: if an object is first perceived in the sequence, then lost, and later seen again, it will be assigned a new instance label unless it can be re-identified as the same object. A static object instance can be tracked in 3D, however, by associating detected instances with the existing portion of the 3D reconstruction even if it leaves the current camera view. 

The data-association task of matching \textit{thing} detections with the reconstruction can be formulated as a Linear Assignment Problem (LAP). Assume we have $n$ detected segments $S_d$, each a set of 3D points associated with an instance label and the segmentation model's confidence scores corresponding to semantic classes. On the other hand, let's assume a set of $m$ tracking targets $S_t$ in the 3D reconstruction formed thus far with similar properties. Since there are no overlaps between the segments in panoptic segmentation and the model's objective is to segment each object as a whole, we can make the common assumption that a detection can only originate from a single target and only one target should be associated with a detection. \cite{ab3dmot,two_stage_data_association,probabilistic_3d_mot} With these assumptions, the problem can be formulated into a $n \times m$ matrix, where each row corresponds to a detection, each column corresponds to a target and cell values are the likelihoods of the the detections originating from from each target. An optimal solution to the problem is then achieved by matching each row with a single column in a way that maximises the sum of likelihoods across all matches.

Because the amount of possible objects in the reconstructed scene is unknown beforehand, a mechanism for generating new target segments is required. With an LAP such a mechanism is quite simple to implement: if an optimal detection-target match has a likelihood below a certain threshold, the detection is assumed to be a new object and it is associated with a new instance label. Since the number of detections and targets might not be equal, a number of temporary dummy IDs are generated to form a square matrix. The likelihood of a detection or a target -- depending on which set is smaller -- matching these dummies is set to zero. Thus, whenever a detection is matched with one -- because there are no more real targets available -- it will be assigned a new ID. On the other hand, dummy detections matched with any target -- when there were less detections than targets -- can be ignored.

The problem can be solved optimally in $\mathcal{O}(n^3)$ time with the Hungarian Algorithm \cite{hungarian}. Variants of the method have been applied in many multi-object tracking works. While faster, approximate, ways to solve the problem have also been invented \cite{probabilistic_3d_mot,two_stage_data_association,panopticfusion}, the optimal solution can be found fast enough in our case since only objects in the current camera view have to be considered. Both PanopticFusion \cite{panopticfusion} and Voxblox++ \cite{voxblox++} apply a greedy algorithm to the task, which is fast, but will not guarantee an optimal solution unless the likelihood threshold is greater than $0.5$. \cite{panoptic_segmentation}

    \subsection{Integrating panoptic labels into a voxel grid}
        \label{subsec:integration}
        Because we represent the environment with a voxel volume, each target object is a set of voxels. Voxel size is assumed uniform across the whole scene. On the other hand -- while detections are originally point clouds -- they are transformed into sets of voxels in the same grid to make them consistent with the targets, and to reduce the number of points used to represent them. We employ a simple yet effective weighting strategy inspired by PanopticFusion \cite{panopticfusion} to determine a voxel's instance ID. For each timestep $t$ and voxel $v$ in the camera view the weight is computed as
\begin{equation}
W(t,v) =
    \begin{cases}
       W(t-1,v) + w(t, v) & l_d(t, v) = l_\tau(t, v), \\
       W(t-1,v) - w(t, v) & l_d(t, v) \neq l_\tau(t, v) \land W(t-1,v) \geq w(t, v) \\
       w_t(v, d) & l_d(t, v) \neq l_\tau(t, v) \land W(t-1,v) < w(t, v) 
    \end{cases}
\end{equation}
where $l_\tau(t, v) \in L$ is the persistent ID of a target instance currently associated with the voxel, $l_d(t, v) \in L$ a persistent ID of a matched detection segment and $w(t, v)$ the TSDF weight of the voxel. If a weight is reduced significantly -- \textit{i.e.} when $l_d(t, v) \neq l_\tau(t, v) \land W(t-1,v) < w(t, v)$, the last case above -- the voxel's persistent instance ID is reset as $l_\tau(t, v) := l_d(t, v)$. This way, only one ID has to be stored, and one does not need to keep track of all the instances associated with each voxel.

With an approach similar to \cite{panopticfusion}, each time a detected segment is matched with a target, the target's confidence scores are integrated in a probabilistic manner. However, because confidence is hard to define in panoptic segmentation, and EfficientPS therefore does not output confidence scores, each detection's confidence is assumed to be one for the detected class and zero for others. Therefore, the semantic class of a target ID $l \in L$ at timestep $t$ is determined by
\begin{equation}
    class(t,l) = 
    \begin{cases}
        argmax_c(n_c^l / n^l) & max_c(n_c^l / n^l) > \theta \\
        void                  & max_c(n_c^l / n^l) \leq \theta
    \end{cases}
\end{equation}
where $n_c^l$ is the number of times the target has been associated with the semantic class $c$, while $n^l$ total number of associations for the target across all classes. If the score is lower than a given threshold $\theta$, it will be assigned to the $void$ semantic class, representing objects that the segmentation model does not recognise.

Imprecise borders between segments in the image can cause problems when back-projecting them to 3D. For instance, if an object's segmentation surpasses the actual object's borders, a part of the detection can appear behind the object \textit{e.g.} on a wall. An accurate segmentation model can alleviate this issue, but nonetheless effective outlier rejection can make the system more robust to segmentation errors. Both confidence intervals on Gaussian probability distributions and clustering with the DBScan algorithm \cite{dbscan} were found to be effective. However, when the voxel weighting approach introduced in \cite{panopticfusion} was employed, the accuracy gains from both approaches were almost insignificant. Therefore, we believe that with a sufficiently precise segmentation model, simply introducing voxel weights as described above is enough. With noisier detections, however, one could benefit from the more complex approaches as well.
        
    \subsection{Association likelihood estimation}
        \label{subsec:likelihood}
        
The likelihood of a detection matching a target can be evaluated in many ways. PanopticFusion \cite{panopticfusion} applies the Intersection over Union (IoU) metric, popular in both evaluation of object detection and image segmentation works \cite{p_voc,panoptic_segmentation}, as well as estimating overlap in object tracking methods \cite{ab3dmot}. On the other hand, Voxblox++ \cite{voxblox++} and it's recent follower \cite{interactive_3d_scenes} simply count intersecting voxels. Applying statistical distance metrics -- \textit{e.g.} the Mahalanobis distance \cite{mahalanobis} -- have also been proposed, increasing accuracy when tracking dynamic objects. \cite{probabilistic_3d_mot,two_stage_data_association}

We found IoU over visible segments to work best in our case. It is computed by dividing the intersection -- in this case, number of intersecting voxels between two segments -- by union -- the total number of voxels between the two. Only parts of the target segments seen in the current camera view are considered, thus objects being only partially visible should not affect the metric as much. The IoU scores are normalised across detection-target pairs to estimate a probability distribution. To avoid setting a fixed threshold for generating new targets, we instead chose to set the threshold as $1/n$, where $n$ is the number of possible matches. In our tests, this strategy seemed to provide better results than any single fixed threshold. The association algorithm's precision and recall could be further tuned by multiplying the threshold with some constant, however we found that to not be necessary in our case.

The same method could be applied with other likelihood metrics as well. Bhattacharyya distance \cite{bhattacharyya} -- a divergence metric between two probability distributions -- was also considered as a likelihood metric. By representing voxel clusters as multivariate Gaussian distributions, we could also take into account the object's shape and size in addition to overlap. However, we found the metric to be less consistent than IoU with the current system and dataset, most likely because many objects are only partially visible.

\section{Evaluation}
    \label{sec:evaluation}

We evaluate the systems performance on the ScanNet dataset \cite{scannet}. Since PanopticFusion \cite{panopticfusion} is the only similar approach on the dataset, we only report its results as a comparison. While \cite{voxblox++} and  \cite{interactive_3d_scenes} are not evaluated on the dataset, their data-association method is quite similar to the one in PanopticFusion, thus their performance can be roughly estimated with its results as well.

In all of the tests below, only every 10th frame of the RGB-D video feed is processed. The original frame-rate of ScanNet data is $30~Hz$, thus with the new rate we are required to process each frame in $333~ms$ or less to run the algorithm in real time. Processing more frames than this does not seem to increase quality of the results much. On the other hand, the number of points in the panoptic point cloud has a huge effect the amount of computation required, therefore the point cloud resolution is reduced from the original $640 \times 480$ points after segmentation, multiplier depending on the voxel size used. The authors of PanopticFusion did not mention which one of Voxblox's TSDF integrators they applied in the article, thus we apply the 'fast' integrator.

To compare them to the ground truth, the results have to be labelled in the original meshes, therefore all our results are transformed to ScanNet evaluation meshes via an approximate nearest neighbour search implemented in Faiss \cite{faiss}. Panoptic Quality is computed similarly to the 2D metric \cite{panoptic_segmentation}, however IoU is computed over mesh vertices instead of pixels. Panoptic image segmentations are inferred separately, and are read from disk during operation so that they can be re-used in all of the tests. All our tests were run on Intel Core i7-8665u CPU at $1.90~GHz$.



%

     
    \subsection{Data}
        \label{subsec:data}
        
All training and evaluation in this manuscript are performed on the ScanNet \cite{scannet} dataset. It contains 2.5 million RGB-D video frames and ground-truth poses from different indoor scenes. Each scene has both 2D and 3D ground-truth annotations with two stuff classes ('wall' and 'floor'), as well as 18 thing classes of objects commonly found in indoor environments. A test set with hidden ground-truth is provided for evaluation of semantic segmentation and object segmentation both in 2D and 3D, but since no panoptic ground-truth is available for the hidden set, a part of the training data is applied as an open test set in this work.

A subset of 25 000 images -- frames sub-sampled from video sequences approximately every 100 frames -- are provided with 2D ground truth for training image segmentation models, which are used to train the panoptic segmentation model. A part of the training data -- consisting of roughly five percent of the images in the 2D training set -- is applied as a validation set. The randomly picked validation scenes are separated from training scenes: locations present in validation dataset are not found in training data. The same scenes found in 2D validation set are also used in evaluating 3D performance. However, since multiple scenes are sometimes captured from the same locations and the reconstructions are usually quite similar, only one instance of each location is stored for 3D evaluation. Therefore, 3D evaluation is performed on 35 randomly picked unique indoor scenes not found in training data.

    \subsection{Panoptic Quality on ScanNet with an open validation set}
        \label{subsec:scannet_open_test}
        
\begin{table}[t]
\caption[Average 3D Panoptic Quality on open ScanNet validation set.]
{Average 3D Panoptic Quality on open ScanNet validation set. Mean Panoptic Quality (PQ), Segmentation Quality (SQ) and Recongnition Quality (RQ) are reported over all classes, as well as over \textit{things} and \textit{stuff} separately.}
\label{tab:scannet_pq}
\begin{center}

\resizebox{\textwidth}{!}{
\begin{threeparttable}
\setlength\tabcolsep{3pt}
\setlength{\belowcaptionskip}{-10pt}

\begin{tabular}{c c c c | c c c | c c c | c}

       & \multicolumn{3}{c}{all} & \multicolumn{3}{c}{stuff} & \multicolumn{3}{c}{things} \\
\multicolumn{1}{c}{} & PQ & SQ & RQ & PQ & SQ & RQ & PQ & SQ & RQ & framerate (Hz)\tnote{2} \\
 
\multicolumn{10}{l}{} \\

\multicolumn{10}{l}{Results reported in PanopticFusion\tnote{1} \cite{panopticfusion}} \\
\hline
\multicolumn{1}{l|}{offline, with CRF} & 33.5 & 73.0 & 45.3 & 58.4 & 70.7 & 80.9 & 30.8 & 73.3 & 41.3 & - \\
\multicolumn{1}{l|}{online, without CRF}  & 29.7 & \textbf{71.2} & 41.1 & \textbf{56.7} & 69.5 & \textbf{79.6} & 26.7 & \textbf{71.4} & 36.8 & 4.30 \\

\multicolumn{10}{l}{} \\
\multicolumn{10}{l}{Results with our method} \\
\hline
\multicolumn{1}{l|}{Greedy, 2.4 cm voxels}   & 29.6 & 66.0 & 41.4 & 51.1 & 69.5 & 71.4 & 27.2 & 65.6 & 38.0 & 0.32 \\
\multicolumn{1}{l|}{Greedy,  5 cm voxels}    & 20.0 & 54.1 & 28.9 & 46.3 & 64.6 & 68.9 & 17.1 & 53.0 & 24.5 & 2.66 \\
\multicolumn{1}{l|}{Greedy,  10 cm voxels}   & 28.9 & 57.0 & 42.4 & 47.3 & 64.7 & 70.3 & 26.9 & 56.1 & 39.3 & 9.83 \\

\multicolumn{10}{c}{} \\

\multicolumn{1}{l|}{Hungarian, 2.4 cm voxels}     & 33.5 & 65.4 & 47.6 & 53.4 & \textbf{70.0} & 74.3 & 31.3 & 64.9 & 44.6 & 0.33 \\
\multicolumn{1}{l|}{Hungarian,  5 cm voxels}       & \textbf{34.0} & 68.0 & \textbf{47.8} & 52.4 & 69.0 & 74.3 & \textbf{31.9} & 67.9 & \textbf{44.5} & 3.53 \\
\multicolumn{1}{l|}{Hungarian,  10 cm voxels}      & 31.7 & 62.3 & 47.0 & 47.2 & 66.5 & 68.8 & 30.0 & 61.9 & 44.5 & \textbf{11.63}\\
\end{tabular}
\begin{tablenotes}\footnotesize
\item[1] Open validation set of the PanopticFusion paper contains different ScanNet scenes than the validation set of this work.
\item[2] PanopticFusion and our method were evaluated on different hardware, thus framerates should not be directly compared between them. They are nevertheless reported to clarify the effect of reducing image and voxel-grid resolutions.
\end{tablenotes}
\end{threeparttable}}

\end{center}
\end{table}

Table \ref{tab:scannet_pq} compares our method to PanopticFusion on the Panoptic Quality (PQ) metric \cite{panoptic_segmentation}, which is a combination of Segmentation Quality (SQ) and Recognition Quality (RQ). The results have been reported as an average over all classes, as well as over stuff and thing classes separately. The results of the original PanopticFusion \cite{panopticfusion} have been reported both with and without Conditional Random Field (CRF) regularisation. Because the test data and panoptic segmentation model are not the same as reported in \cite{panopticfusion}, we have also re-implemented the Greedy data-association algorithm of PanopticFusion: detection segments are processed in descending order based on size, by choosing the target corresponding to highest IoU. IoU threshold for the greedy algorithm is set to $0.25$ as in the original work. The results on the test closest to the original PanopticFusion results -- Greedy algorithm with $2.4~cm$ voxels -- seem to be quite close to the results reported in the original paper.

\begin{figure}[t]
\setlength{\belowcaptionskip}{-10pt}
\centering
\begin{subfigure}[b]{0.32\textwidth}  
        \includegraphics[width=\textwidth]{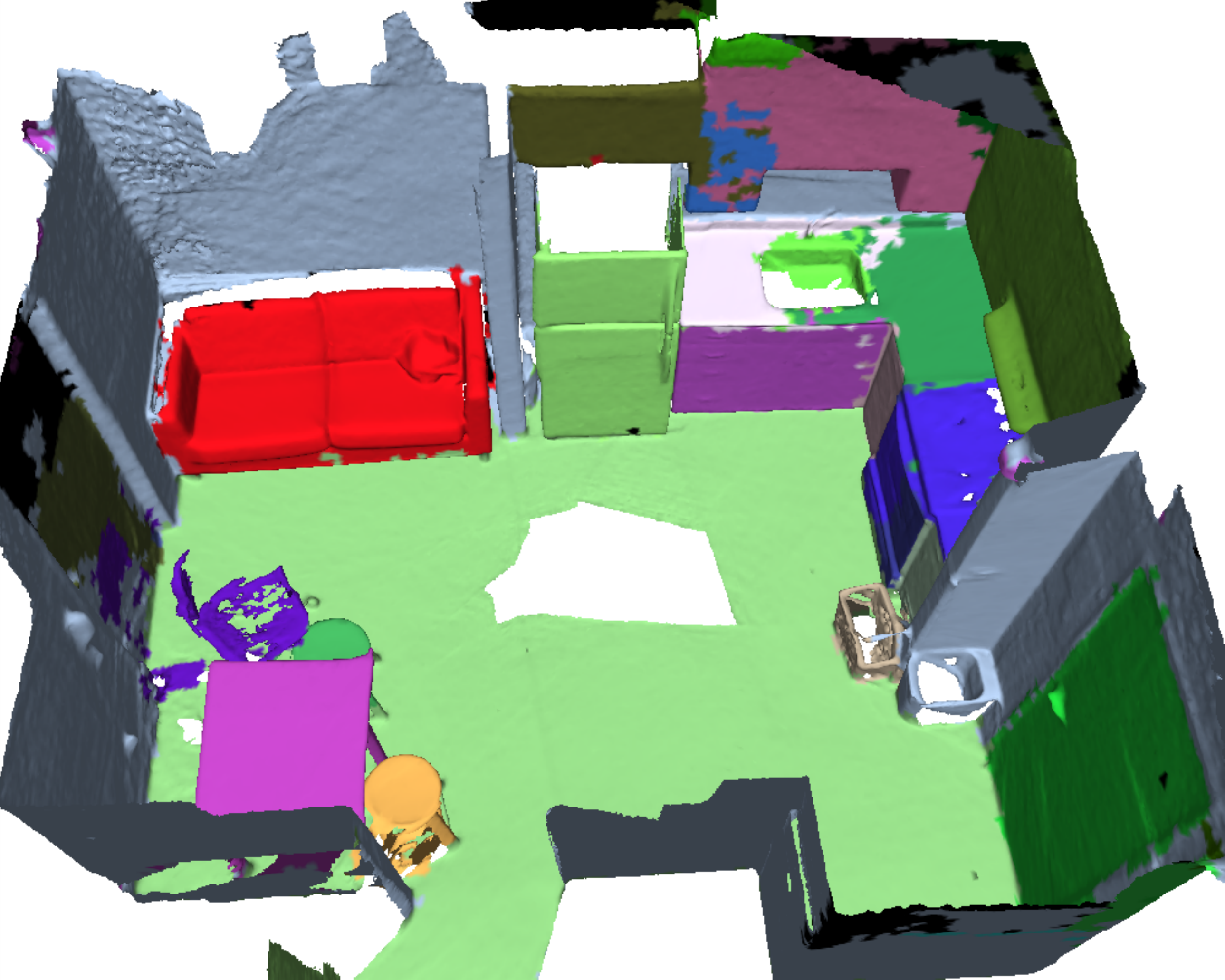}
\end{subfigure}
\begin{subfigure}[b]{0.32\textwidth}  
        \includegraphics[width=\textwidth]{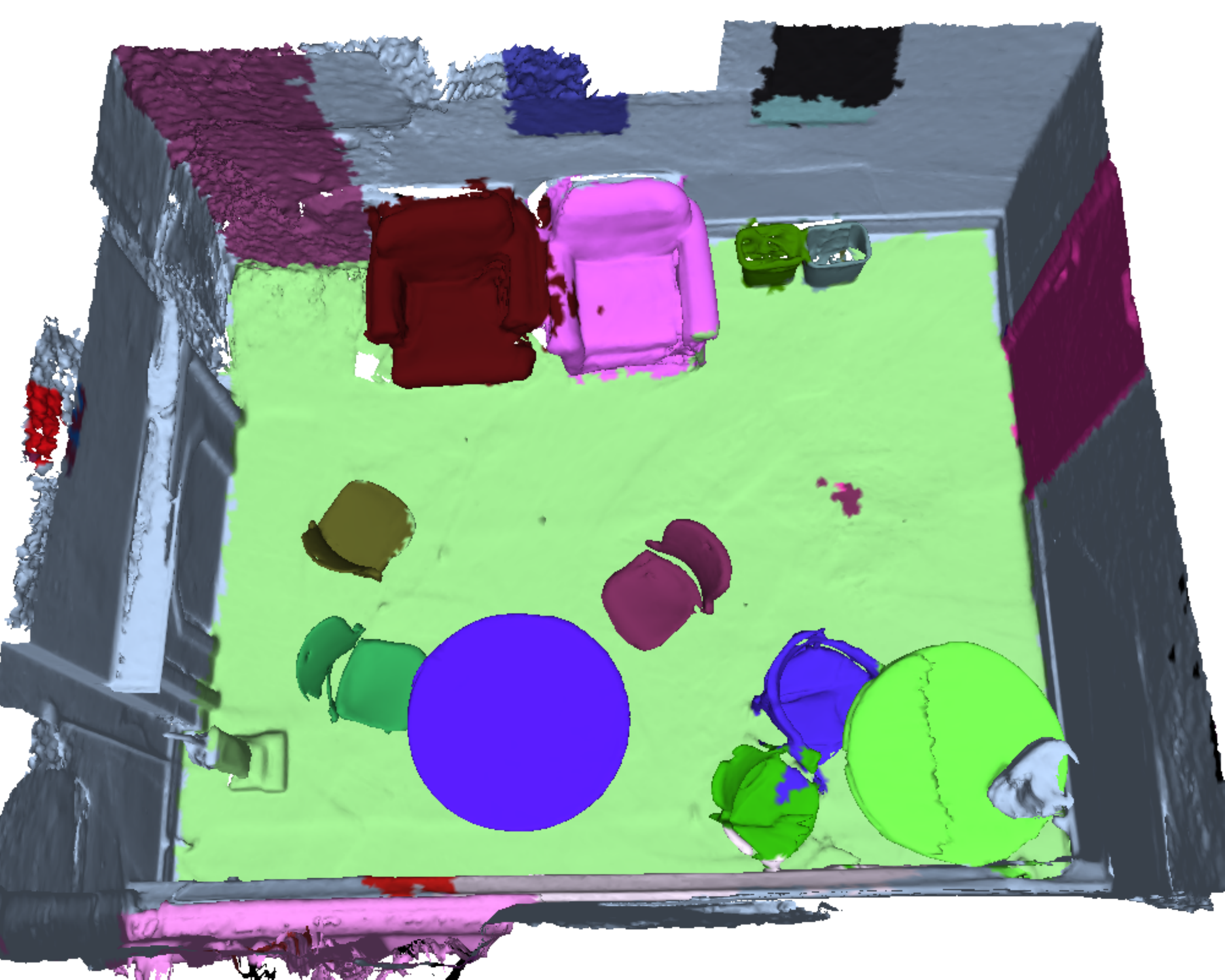}
\end{subfigure}
\begin{subfigure}[b]{0.32\textwidth}  
        \includegraphics[width=\textwidth]{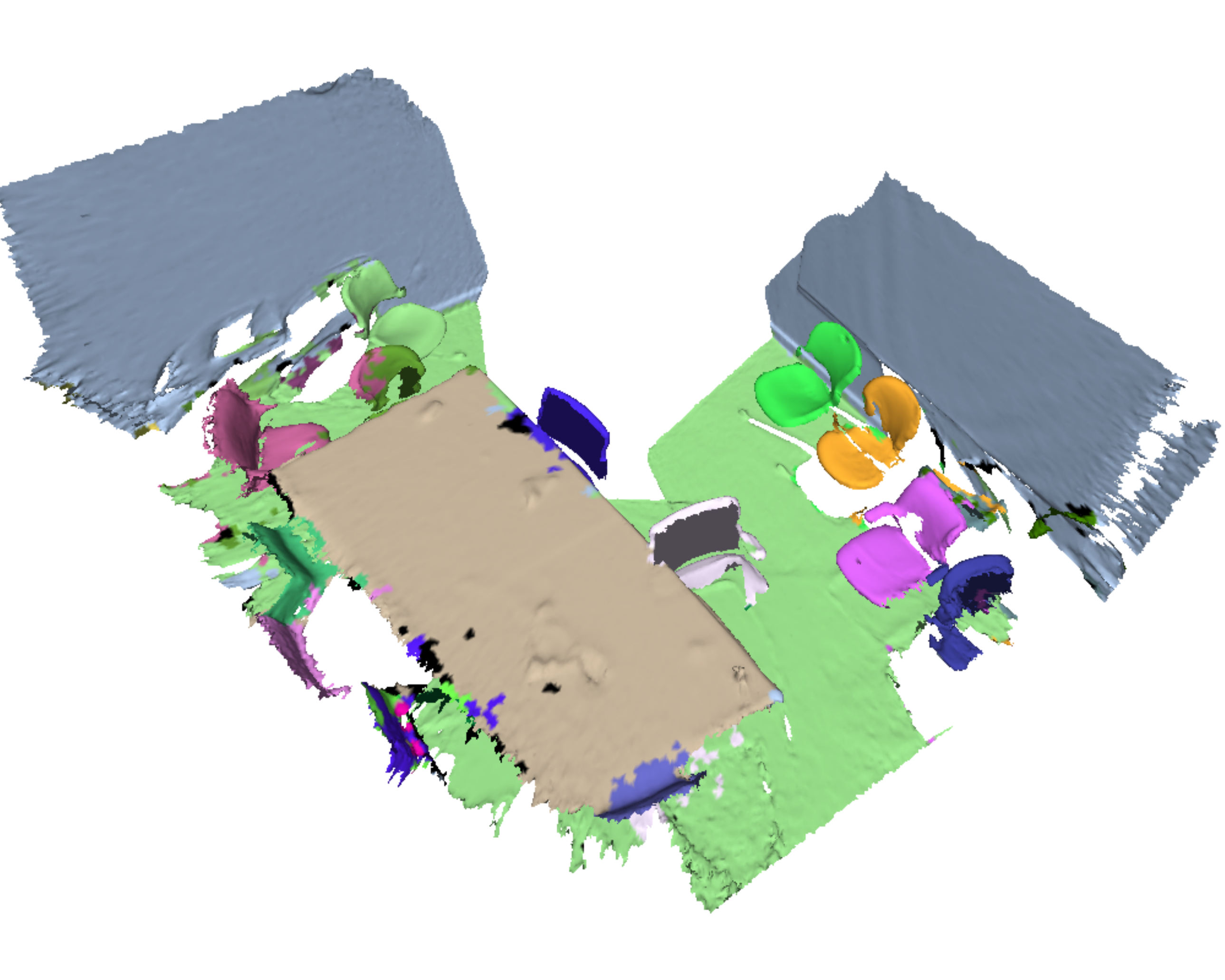}
\end{subfigure}
\hfill
\begin{subfigure}[b]{0.32\textwidth}  
        \centering 
        \includegraphics[width=\textwidth]{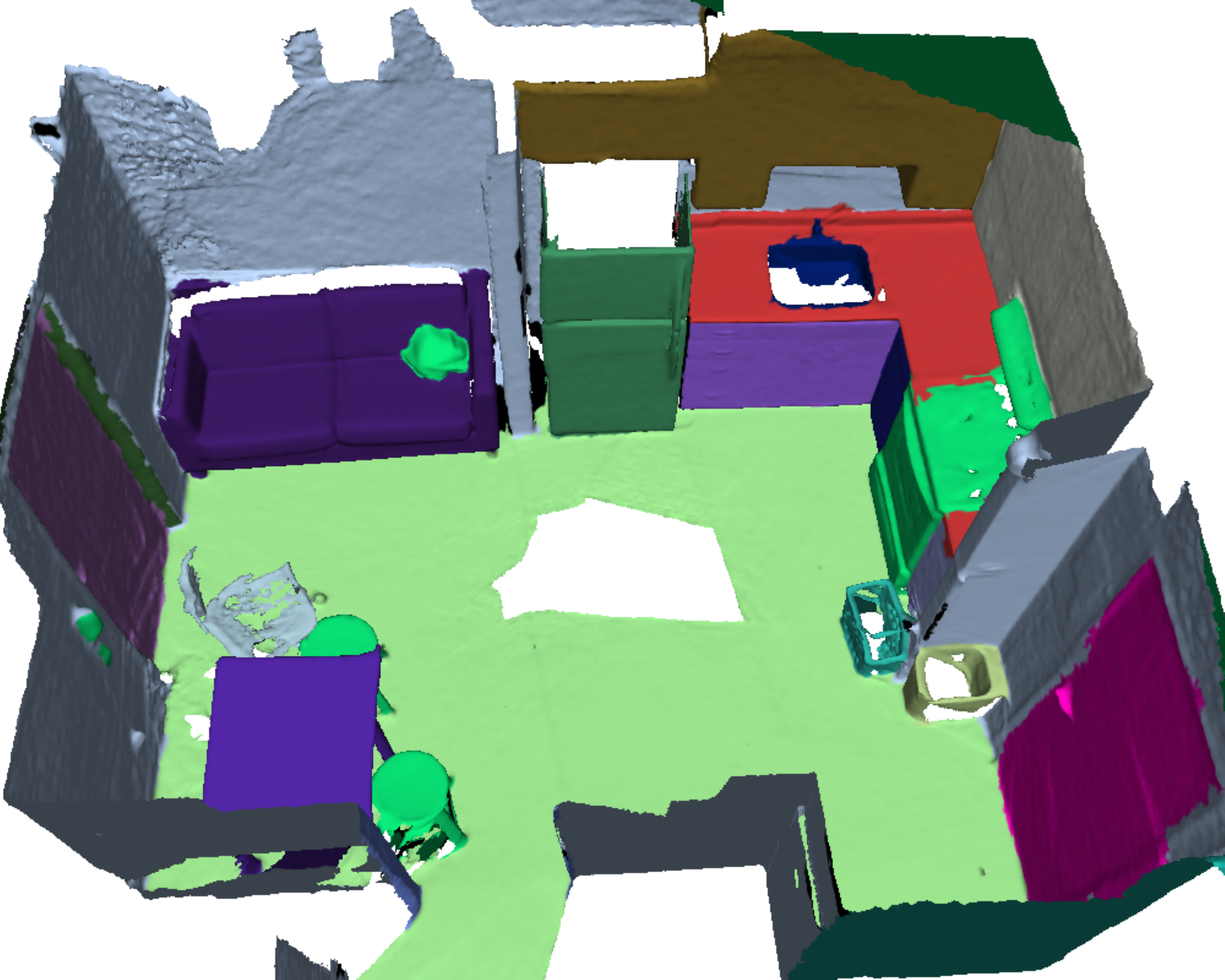}
        \caption[]%
        {\small scene $0288\_00$}
        \label{fig:288}
\end{subfigure}
\begin{subfigure}[b]{0.32\textwidth}  
        \centering 
        \includegraphics[width=\textwidth]{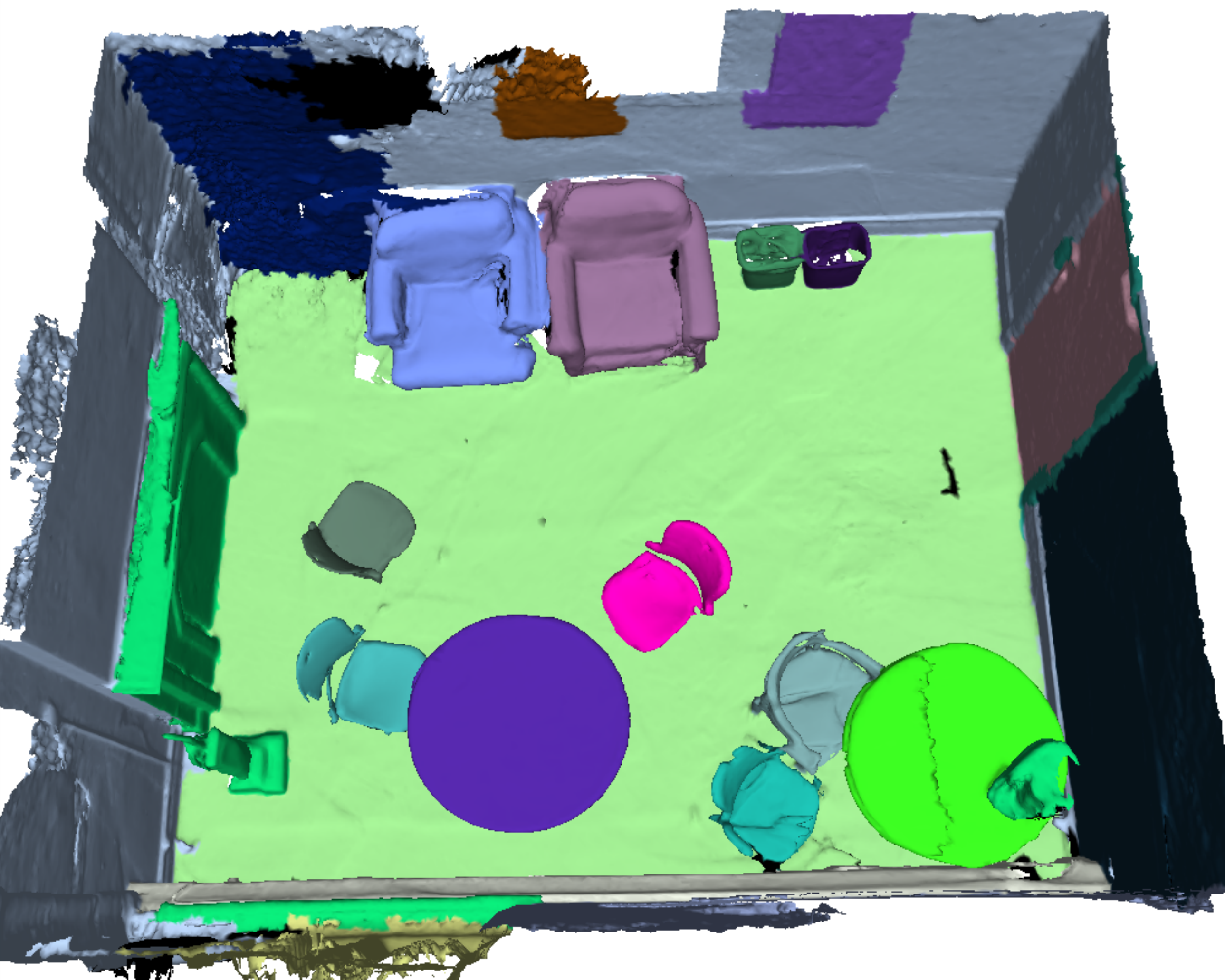}
        \caption[]%
        {\small scene $0616\_00$}    
        \label{fig:616}
\end{subfigure}
\begin{subfigure}[b]{0.32\textwidth}  
        \centering 
        \includegraphics[width=\textwidth]{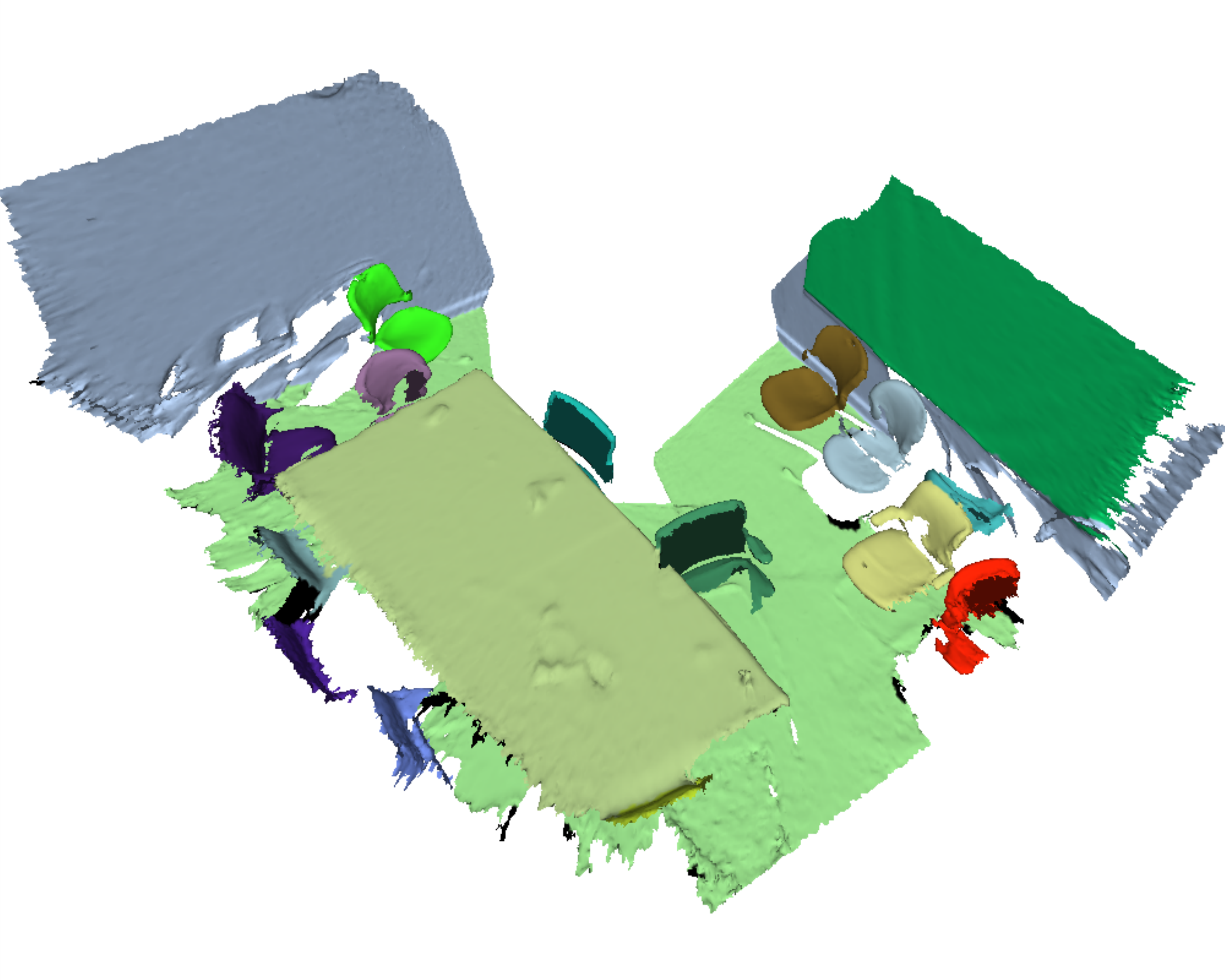}
        \caption[]%
        {\small scene $0655\_00$}  
        \label{fig:655}
\end{subfigure}
\caption[]%
{Qualitative examples of typical scenes in the ScanNet dataset. Top: panoptic reconstructions with our method, bottom: ground truth. Colors of 'thing' instances can be different from ground truth because they are randomised.}    
\label{fig:examples}
\end{figure}

The visual quality of the reconstruction varies mostly based on the resolution of the voxel grid: the smaller the voxels are, the more detailed the meshes become. Voxel size seems to affect computational requirements the most as well. While image resolution could be reduced up to $0.1$ times the original after segmentation with voxels larger than $5~cm$ without affecting the quality of the outcome much, with smaller voxels there is a significant loss of quality. When image resolution is only reduced to half of the original, the quality is much better, but computational requirements grow respectively. In tests with $5~cm$ and $10~cm$ voxels point cloud resolution was reduced to $0.1$ times the original, and in tests with $2.4~cm$ voxels the resolution was reduced to $0.5$ times the original.

While our method seems to be roughly on par with the version of PanopticFusion applying CRF, the one without CRF seems to perform significantly worse than ours on PQ. This is further supported by the tests on our re-implementation of their method. While both PanopticFusion configurations perform better on 'stuff' classes and Segmentation Quality (SQ) than ours, they both seem to fall behind on 'thing' classes and Recognition Quality (RQ). This implies that while our method's segmentation seems to be slightly worse, our data association strategy seems to be performing better. Differences in SQ most likely originate from differences in the image segmentation approaches.

Some examples of panoptic reconstruction on typical scenes found in ScanNet are presented in Figure \ref{fig:examples}. Object instances seem to have been found and separated from one another correctly most of the time, and segmentations look pretty accurate, even though larger objects are often only seen partially in the video sequences.

Volumetric integration slows down a lot when voxel size is decreased taking on average $46$, $106$ or $1188$ milliseconds for $10$, $5$ and $2.4$ centimetre voxels respectively. Data-association with the Hungarian algorithm took on average $47$, $177$ or $1875$ milliseconds for the same sizes, while greedy matching took on average $53$, $247$ or $1949$ milliseconds. Most of this time is spent on computing the IoUs, which takes more time with larger amounts of voxels. Compared to the relations of these timings reported in the PanopticFusion paper, our association algorithm is somewhat slower than the original. Our implementation of the greedy algorithm is a bit slower than the Hungarian method, most likely because the segments need to be sorted every iteration. We can reach reconstruction quality similar to the version of PanopticFusion with CRF regularisation even when decimating point clouds, reducing voxel grid resolution and processing less frames. While the CRF is only run every $10$ seconds, it is reported to take around $4.5$ seconds and to slow down along increasing reconstruction size. For a real-time application to get the benefits of the regularisation, one would need to tolerate long delays during operation, thus we see it more useful as a post-processing tool for cleaning the reconstruction offline.

\section{Conclusion}
    \label{sec:conclusion}
    


In this work, we revisited the idea of sequentially integrating panoptic image segmentation to 3D reconstruction introduced in \cite{panopticfusion}. We formulated the task as a Linear Assignment Problem and studied a way of solving it optimally fast enough for real-time applications. Our method seems to outperform earlier works when operating strictly online.

In the future, we aim to research real-time applications that benefit from the more sophisticated scene understanding that panoptic reconstruction offers. Other possible research topics also include extending this method for tracking dynamic objects in 3D simultaneously and applying the method for different sensor modalities. Because input data is only required to be segmented point clouds and pose information, the system could also be adapted for LiDARs and multi-camera setups relatively easily. These would provide increased real-time situational awareness compared to a single camera, albeit at the cost of increased computational requirements. Because the method only processes data seen in the current camera view, it should also be scalable to larger environments as well.

\bibliographystyle{splncs04}
\bibliography{refs}

\begin{thebibliography}{10}
\providecommand{\url}[1]{\texttt{#1}}
\providecommand{\urlprefix}{URL }
\providecommand{\doi}[1]{https://doi.org/#1}

\bibitem{s3dis}
Armeni, I., Sener, O., Zamir, A.R., Jiang, H., Brilakis, I., Fischer, M.,
  Savarese, S.: 3d semantic parsing of large-scale indoor spaces. IEEE/CVF
  Conference on Computer Vision and Pattern Recognition (CVPR)  (2016)

\bibitem{bhattacharyya}
Bhattacharyya, A.: On a measure of divergence between two multinomial
  populations. Sankhyā: The Indian Journal of Statistics  \textbf{7}(4),
  401--406 (1946)

\bibitem{panoptic_deeplab}
Cheng, B., Collins, M.D., Zhu, Y., Liu, T., Huang, T.S., Adam, H., Chen, L.C.:
  Panoptic-deeplab: A simple, strong, and fast baseline for bottom-up panoptic
  segmentation. IEEE/CVF Conference on Computer Vision and Pattern Recognition
  (CVPR)  (2020)

\bibitem{probabilistic_3d_mot}
Chiu, H.k., Prioletti, A., Li, J., Bohg, J.: Probabilistic 3d multi-object
  tracking for autonomous driving. arXiv preprint  (2020)

\bibitem{tsdf}
Curless, B., Levoy, M.: A volumetric method for building complex models from
  range images. ACM SIGGRAPH  (1996)

\bibitem{scannet}
Dai, A., Chang, A.X., Savva, M., Halber, M., Funkhouser, T., Nie{\ss}ner, M.:
  Scannet: Richly-annotated 3d reconstructions of indoor scenes. IEEE/CVF
  Conference on Computer Vision and Pattern Recognition (CVPR)  (2017)

\bibitem{two_stage_data_association}
Dao, M.Q., Frémont, V.: A two-stage data association approach for 3d
  multi-object tracking. MDPI Sensors  \textbf{21}(9) (2021)

\bibitem{dbscan}
Ester, M., Kriegel, H.P., Sander, J., Xu, X.: A density-based algorithm for
  discovering clusters in large spatial databases with noise. International
  Conference on Knowledge Discovery and Data Mining p. 226–231 (1996)

\bibitem{p_voc}
Everingham, M., Gool, L.V., Williams, C.K.I., Winn, J.M., Zisserman, A.: The
  pascal visual object classes (voc) challenge. International Journal of
  Computer Vision  \textbf{88},  303--338 (2009)

\bibitem{voxblox++}
{Grinvald}, M., {Furrer}, F., {Novkovic}, T., {Chung}, J.J., {Cadena}, C.,
  {Siegwart}, R., {Nieto}, J.: {Volumetric Instance-Aware Semantic Mapping and
  3D Object Discovery}. IEEE Robotics and Automation Letters  \textbf{4}(3),
  3037--3044 (2019)

\bibitem{interactive_3d_scenes}
Han, M., Zhang, Z., Jiao, Z., Xie, X., Zhu, Y., Zhu, S.C., Liu, H.:
  Reconstructing interactive 3d scenes by panoptic mapping and cad model
  alignments. IEEE International Conference on Robotics and Automation (ICRA)
  (2021)

\bibitem{panoptic_mope}
Hoang, D.C., Lilienthal, A.J., Stoyanov, T.: Panoptic 3d mapping and object
  pose estimation using adaptively weighted semantic information. IEEE Robotics
  and Automation Letters  \textbf{5}(2),  1962--1969 (2020)

\bibitem{real_time_panoptic}
Hou, R., Li, J., Bhargava, A., Raventos, A., Guizilini, V., Fang, C., Lynch,
  J., Gaidon, A.: Real-time panoptic segmentation from dense detections.
  IEEE/CVF Conference on Computer Vision and Pattern Recognition (CVPR)  (2020)

\bibitem{faiss}
Johnson, J., Douze, M., J{\'e}gou, H.: Billion-scale similarity search with
  gpus. arXiv preprint arXiv:1702.08734  (2017)

\bibitem{video_panoptic_segmentation}
Kim, D., Woo, S., Lee, J.Y., Kweon, I.S.: Video panoptic segmentation. IEEE/CVF
  Conference on Computer Vision and Pattern Recognition (CVPR)  (2020)

\bibitem{panopticfpn}
Kirillov, A., Girshick, R., He, K., Dollar, P.: Panoptic feature pyramid
  networks. IEEE/CVF Conference on Computer Vision and Pattern Recognition
  (CVPR)  (2019)

\bibitem{panoptic_segmentation}
Kirillov, A., He, K., Girshick, R., Rother, C., Dollar, P.: Panoptic
  segmentation. IEEE/CVF Conference on Computer Vision and Pattern Recognition
  (CVPR)  (6 2019)

\bibitem{hungarian}
Kuhn, H.: The hungarian method for the assignment problem. Naval Research
  Logistic Quarterly  \textbf{2} (1955)

\bibitem{mahalanobis}
Mahalanobis, P.C.: On the generalized distance in statistics. Proceedings of
  the National Institute of Sciences of India  (1936)

\bibitem{efficientps}
Mohan, R., Valada, A.: Efficientps: Efficient panoptic segmentation.
  International Journal of Computer Vision (IJCV)  (2021)

\bibitem{panopticfusion}
Narita, G., Seno, T., Ishikawa, T., Kaji, Y.: Panopticfusion: Online volumetric
  semantic mapping at the level of stuff and things. IEEE/RSJ International
  Conference on Intelligent Robots and Systems (IROS) pp. 4205--4212 (2019)

\bibitem{voxblox}
Oleynikova, H., Taylor, Z., Fehr, M., Siegwart, R., Nieto, J.: Voxblox:
  Incremental 3d euclidean signed distance fields for on-board mav planning.
  IEEE/RSJ International Conference on Intelligent Robots and Systems (IROS)
  (2017)

\bibitem{seamless}
Porzi, L., Rota~Bul\`o, S., Colovic, A., Kontschieder, P.: Seamless scene
  segmentation. IEEE/CVF Conference on Computer Vision and Pattern Recognition
  (CVPR)  (2019)

\bibitem{paris_lille_3d}
Roynard, X., Deschaud, J.E., Goulette, F.: Paris-lille-3d: A large and
  high-quality ground-truth urban point cloud dataset for automatic
  segmentation and classification. The International Journal of Robotics
  Research  \textbf{37}(6) (2018)

\bibitem{ab3dmot}
Weng, X., Wang, J., Held, D., Kitani, K.: {3D Multi-Object Tracking: A Baseline
  and New Evaluation Metrics}. IROS  (2020)

\bibitem{scenegraphfusion}
Wu, S.C., Wald, J., Tateno, K., Navab, N., Tombari, F.: Scenegraphfusion:
  Incremental 3d scene graph prediction from rgb-d sequences. In: IEEE/CVF
  Conference on Computer Vision and Pattern Recognition (CVPR). pp. 7515--7525
  (2021)

\end{thebibliography}

\end{document}